\documentclass{bmvc2k}
\usepackage{varwidth}
\usepackage{float}
\graphicspath{{figures/}}
\usepackage{xcolor}

\newcommand{\Eq}[1]{Eq.~(\ref{eq:#1})}
\newcommand{\eq}[1]{\Eq{#1}}
\newcommand{\fig}[1]{Fig.~\ref{fig:#1}}
\newcommand{\tab}[1]{Tab.~\ref{tab:#1}}

\usepackage{graphicx}
\usepackage{amsmath}
\usepackage{amssymb}
\usepackage{booktabs}
\usepackage{comment}
\usepackage{pifont}
\newcommand{\cmark}{\ding{51}}%
\newcommand{\xmark}{\ding{55}}%

\title{NeRD++: Improved 3D-mirror symmetry learning from a single image}
\addauthor{Yancong Lin\\ Silvia L. Pintea \\Jan  C. van Gemert} {}{}

\addinstitution{
Computer Vision Lab,\\
Delft University of Technology,\\
Delft, Netherlands\\
}
\runninghead{\scriptsize Y. Lin \etal}{\scriptsize Improved 3D-mirror symmetry learning from a single image}

\def\etal{\emph{et al}\bmvaOneDot}

\begin{document}
\maketitle

\begin{abstract}
Many objects are naturally symmetric, and this symmetry can be exploited to infer unseen 3D properties from a single 2D image. Recently, NeRD \cite{zhou2021nerd} is proposed for accurate 3D mirror plane estimation from a single image. Despite the unprecedented accuracy, it relies on large annotated datasets for training and suffers from slow inference. Here we aim to improve its data and compute efficiency.
We do away with the computationally expensive 4D feature volumes and instead explicitly compute the feature correlation of the pixel correspondences across depth, thus creating a compact 3D volume. We also design multi-stage spherical convolutions to identify the optimal mirror plane on the hemisphere, whose inductive bias offers gains in data-efficiency. 
Experiments on both synthetic and real-world datasets show the benefit of our proposed changes for improved data efficiency and inference speed.

\end{abstract}

\section{Introduction}
Symmetry exists in nature, in man-made environments, in science and arts. 
Mirror symmetry, also known as bilateral or reflection symmetry, is an intrinsic property of many objects, and it allows infering the entire object from only partial view.
This has been shown to be useful for shape completion \cite{gao2020psrnet, liu2020morphing} and single-view 3D reconstruction \cite{yao2020front2back, xu2019disn, wu2020unsupervised}. 


Deep learning approaches have achieved astonishing results on estimating mirror symmetries from single-view images, by learning dense features from convolutional networks instead of relying on local feature matching \cite{gao2020psrnet, shi_siga20, zhou2021nerd}. In addition to feature learning, deep networks can incorporate 3D mirror geometry as in NeRD \cite{zhou2021nerd}, the top-performing symmetry detection model. 
However, NeRD \cite{zhou2021nerd} builds a computationally 4D feature volumes during learning, resulting in high inference latency. 
Moreover, its performance also deteriorates substantially when limited training data is available.

\begin{figure}[!t]
    \centering
    \includegraphics[width=0.8\textwidth]{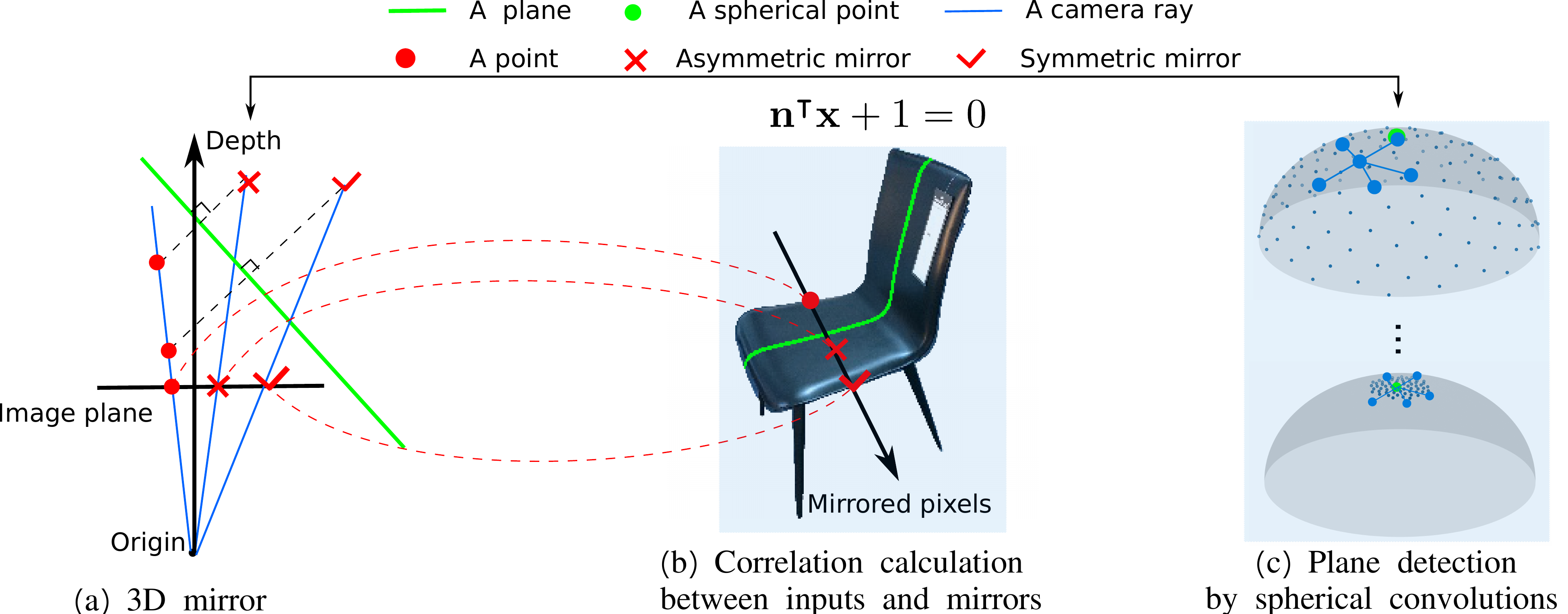}
    \caption{\small
    \textbf{3D mirror symmetry detection.} 
    We identify 3D mirror planes by measuring correlations between the input and its mirrors over depth.
    The mirrors are computed by 3D mirror geometry (a) as in NeRD \cite{zhou2021nerd}, which localizes the reflections of a given pixel directly on the image plane (b). 
    We measure the similarity between a pixel and its mirrors by explicitly calculating correlations which indicate to what extent a point resembles its correspondences.
    We also adopt multi-stage spherical convolutions to localize the optimal plane hierarchically on the hemisphere where all candidate planes are sampled (c). 
    We improve both data and compute efficiency over NeRD \cite{zhou2021nerd} by introducing (b) and (c).
    }
    \label{fig:mirror_exp}
\end{figure}

In this paper, we make two improvements over NeRD \cite{zhou2021nerd} to increase its data and compute efficiency. 
Specifically, from learned semantic deep features we calculate a compact 3D correlation volume for each candidate plane.
Correlations measure the similarity between the features and their mirrored versions, at every location in the featuremap. 
The optimal plane is characterized by the highest feature correlation.
By adding explicit correlation computation into the model, we bypass the expensive 4D feature volumes in NeRD \cite{zhou2021nerd}, and thus substantially speed up the inference. 
Our second modification is on data-efficiency by exploiting geometric priors~\cite{lin2020deep,lin2022deep} where we use spherical convolutions on the hemisphere.
The hemisphere is the space containing all the candidate symmetry planes. 
Rather than using a huge fully connected layer as in NeRD to locate the optimal symmetry plane, we make use of the geometric prior that the shape of the search-space is a hemisphere: We use spherical convolutions as a prior~\cite{lin2022deep}. 
These two well-chosen inductive biases, are what contributes to both the data-efficiency and computational efficiency of our model.
\fig{mirror_exp} summarizes our approach.


Our contributions are: (1) improving the data efficiency of NeRD, the top-performing method on 3D mirror symmetry detection from single images by introducing spherical convolutions; (2) reducing the inference latency significantly ($\times$ 20), by calculating explicit correlations; (3) experimentally demonstrating the added-value of our improvements on two datasets: ShapeNet \cite{chang2015shapenet} and Pix3D \cite{sun2018pix3d}.

\section{Related work}
\smallskip\noindent\textbf{Planar symmetry detection.} 
A thorough overview on symmetry detection with focus on 2D symmetries, is given in \cite{liu2013symmetry}.
Further work expands on this by including other types of symmetries such as medial-axis-like symmetries and by adding synthetic 3D data \cite{funk20172017}.
More recently, planar symmetry detection with deep networks achieves competitive results \cite{funk2017beyond, seo2021learning}.
However, for planar symmetry detection objects are typically front-facing, greatly simplifying the task. 
In addition, planar symmetry does not encode any 3D perspective information. 
Here we differ from these works, as we focus on 3D mirror symmetry from single-view images taken from any perspective. 

\smallskip\noindent\textbf{3D mirror symmetry detection.} 
3D mirror symmetry is prevalent in both nature and the man-made environments. 
There has been excellent research on utilizing geometric transforms for detecting mirror symmetries from 3D inputs \cite{cailliere20083d, sfikas2011rosy, podolak2006planar}. 
A 3D Hough transform proves effective at detecting mirror symmetry planes from point clouds, in \cite{cailliere20083d}. 
Alternatively, planar reflective symmetry transform can find symmetry planes in 3D volumetric data \cite{podolak2006planar}.
Similarly, we also focus on 3D geometric priors, and specifically on how to improve existing priors, but instead of relying on 3D data we start from single-view images. 

Recently, deep networks have been used for leveraging large datasets for learning 3D symmetries \cite{shi_siga20,gao2020psrnet}. 
Despite being able to detect multiple symmetries, they rely on heavy post-processing procedures to find the optimal symmetry plane. 
Moreover, these models have only been tested on synthetic 3D datasets with voxelized volumes or RGB-D data (i.e. ShapeNet \cite{chang2015shapenet}). 
In contrast, we propose improvements for end-to-end 3D mirror symmetry detection, and test on both synthetic and real-world 2D images.

\smallskip\noindent\textbf{3D mirror symmetry from single-view images.} 
A 2-stage approach can be effective for 3D mirror symmetry detection from 2D images, by first matching image correspondences and then applying RANSAC to identify the best symmetry plane \cite{koser2011dense}. 
However, this strategy is no longer applicable in the absence of texture, or on smooth surfaces, or repetitive patterns, because of incorrect correspondences. 
Rather than relying on local feature matching, NeRD \cite{zhou2021nerd} makes use of neural networks to learn dense features, making it the top-performing model. 
Here, we use NeRD \cite{zhou2021nerd} as our starting point and make two essential changes to the model: we explicitly compute correlations between correspondences and we use spheric convolutions to correctly localize the optimal symmetry plane on the hemisphere. These two extensions greatly reduce computations and make our model data-efficient. 

\section{Revisiting 3D mirror geometry and NeRD} \label{background}
\subsection{3D mirror geometry}

\begin{figure}
\centering
\begin{tabular}{c@{\hskip 0.5in}c}
    \includegraphics[width=.4\textwidth]{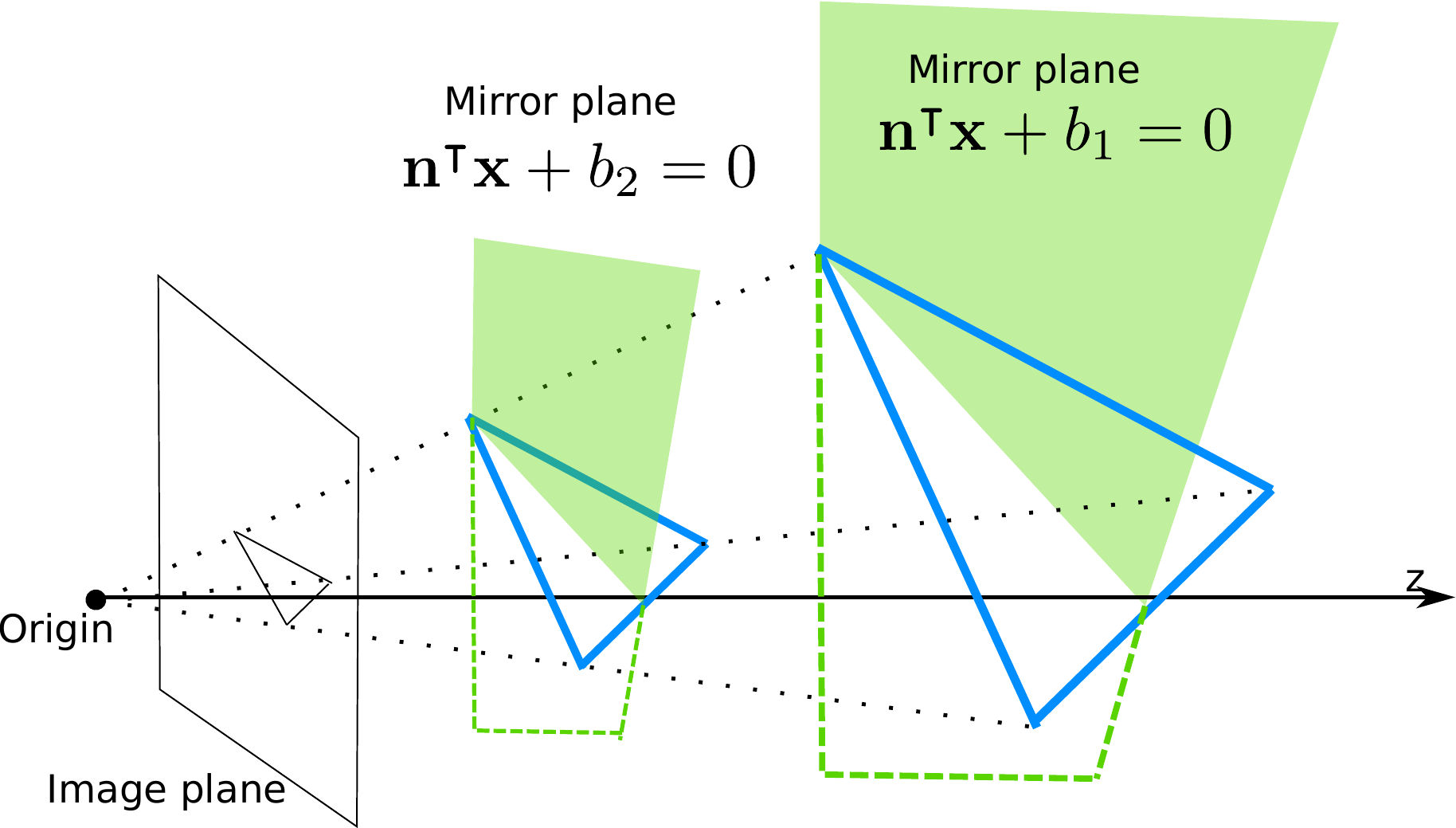} &
    \includegraphics[width=.3\textwidth]{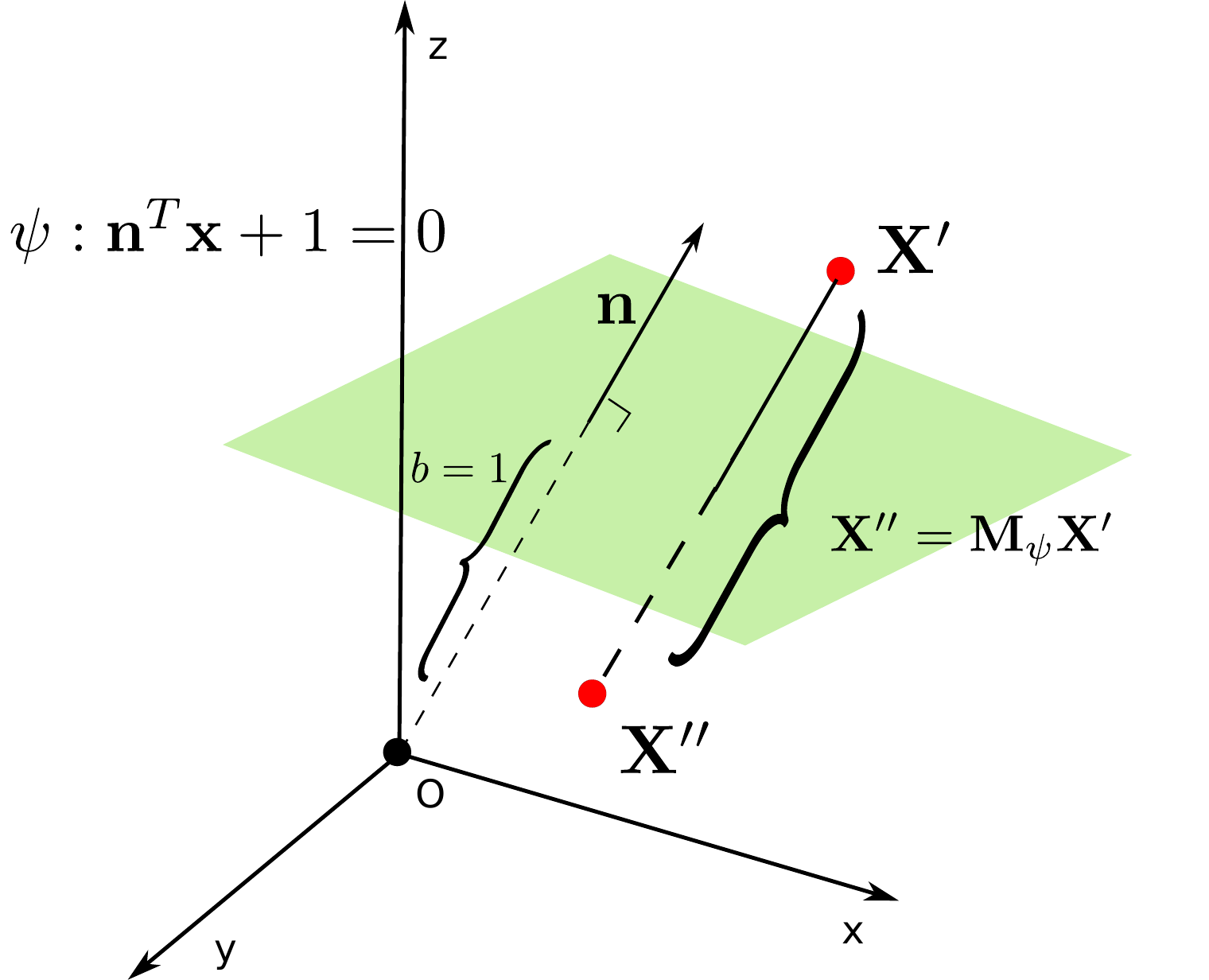}\\
    \small{(a) Scale ambiguity} & \small{(b) 3D mirror geometry} \\
    \end{tabular}
\caption{\small
    \textbf{(a) Scale ambiguity.} 
    The two objects (in blue) only differ in scale, but their projections on the image plane are the same. 
    Therefore, we are unable to determine the scale of the object, or the value of the plane offset. 
    In practice, we set $b=1$. 
    \textbf{(b) 3D mirror geometry.} 
    3D points $\mathbf{X^\prime}$ and $\mathbf{X^{\prime\prime}}$ are symmetric with respect to the given plane $\psi$ defined by $\mathbf{n}^\intercal \mathbf{x} + 1 = 0$. 
    We represent $\mathbf{X^\prime} \in \mathbb{R}^{4}$ and $\mathbf{X^{\prime\prime}} \in \mathbb{R}^{4}$ in the homogeneous coordinate, where $\mathbf{X^{\prime\prime}} = \mathbf{M_{\psi}} \mathbf{X^\prime} $ as defined in \eq{symmetry}. 
    $\mathbf{M_{\psi}}\in \mathbb{R}^{4 \times 4}$ is the 3D mirror transformation uniquely determined by the normal direction $\mathbf{n}$ \cite{cailliere20083d, koser2011dense}. 
}
\label{fig:symmetry}
\end{figure}

\medskip\noindent\textbf{(i) 3D mirror planes.}
A plane $\psi$ is uniquely defined by its normal direction $\mathbf{n} \in \mathbb{R}^{3}$ and offset $b \in \mathbb{R}$ as $\mathbf{n}^\intercal \mathbf{x} + b = 0$, where $\mathbf{x}\in \mathbb{R}^{3} $ denotes points on the plane. 
However, we are unable to determine $b$ from a single image due to scale ambiguity \cite{ma2012invitation, zhou2021nerd}. 
This is because the scene can be moved arbitrarily along the normal direction $\mathbf{n}$ and scaled accordingly, without affecting the image, as shown in \fig{symmetry}(a). 
Therefore, $b$ is often set to 1 and the normal direction $\mathbf{n}$ of the mirror plane is only the unknown to predict. 
Moreover, given that a normal direction $\mathbf{n}$ is equivalent to a point on a unit hemisphere, we can further define a plane as a spherical point. 
Thus we can sample planes on a unit hemisphere. 

\medskip\noindent\textbf{(ii) 3D mirror transform.} 
\fig{symmetry}(b) shows an illustration of 3D mirror geometry for a randomly sampled plane $\psi$ defined by $ \mathbf{n}^\intercal \mathbf{x} + 1 = 0 $. 
The corresponding 3D mirror transformation $\mathbf{M_{\psi}} \in \mathbb{R}^{4 \times 4}$ associated to plane $\psi: \mathbf{n}^\intercal \mathbf{x} + 1 = 0$ is uniquely defined by the normal direction of the plane $\mathbf{n}$ \cite{cailliere20083d, koser2011dense}: 
\begin{equation}
\mathbf{X^{\prime\prime}}=
\underbrace{
\begin{pmatrix}
\mathbf{I} -2\mathbf{n}\mathbf{n}^\intercal & -2 \mathbf{n}\\
\mathbf{0} & 1 
\end{pmatrix}}_{\mathbf{M_{\psi}}}
\mathbf{X^{\prime}},
\label{eq:symmetry}
\end{equation}
where $\mathbf{X^{\prime\prime}}\in \mathbb{S}$ and $\mathbf{X^{\prime}}\in \mathbb{S}$ are a pair of symmetric 3D points, and $\mathbb{S}\subset \mathbb{R}^{4}$ is the set of 3D points on the object surface in homogeneous coordinates.

Given the camera intrinsic matrix $\mathbf{K}\in \mathbb{R}^{4 \times 4}$, both $\mathbf{X^{\prime}}$ and $\mathbf{X^{\prime\prime}}$ can be projected on the image plane by $\mathbf{x^{\prime}} = \mathbf{K} \mathbf{X^{\prime}} /d^{\prime}$ and $\mathbf{x^{\prime\prime}} = \mathbf{K} \mathbf{X^{\prime\prime}} /d^{\prime\prime}$, where $d^{\prime}$ and $d^{\prime\prime}$ are the corresponding depths in the camera space. 
Thus, the constraint between points $\mathbf{x^{\prime}}$ and their projections $\mathbf{x^{\prime\prime}}$ can be derived as:  
\begin{equation}
\mathbf{x^{\prime\prime}} d^{\prime\prime} =
\mathbf{K} \mathbf{M_{P}} \mathbf{K}^{-1}\mathbf{x^{\prime}} d^{\prime},
\label{eq:symmetry2}
\end{equation}
where $\mathbf{x^{\prime}} = \left [x^{\prime}, y^{\prime}, 1, 1/d^{\prime} \right ]$ and $\mathbf{x^{\prime\prime}} = \left [x^{\prime\prime}, y^{\prime\prime}, 1, 1/d^{\prime\prime} \right ]$ indicate the coordinates of the projected points in the pixel space. 
\eq{symmetry2} enables us to find the symmetric correspondences of every pixel at various depths, given a sampled mirror plane, $\psi$.

\subsection{NeRD overview}
We briefly recap NeRD, the state-of-the-art model on symmetry detection, which incorporates 3D mirror geometry into learning.
NeRD represents our starting point.

\medskip\noindent\textbf{(i) Feature learning.}
NeRD first learns semantic features from RGBA images via a convolutional neural network, which results in a feature map $\mathcal{F}$ of size [$H \times W \times C$], where $H$, $W$, $C$ indicate height, width, and number of channels, respectively.

\medskip\noindent\textbf{(ii) 3D mirroring.}
Given a randomly sampled plane from the hemisphere, $\psi$, NeRD localizes the symmetric correspondences $(x^{\prime},y^{\prime})$ for each pixel $(x,y)$ at varying depth $d \in \mathcal{D}$ according to \eq{symmetry2}, where $\mathcal{D}=\{d_{min}+ \frac{i}{D-1} (d_{max}-d_{min}) \lvert i=\{0,1,...,D-1\}\}$,  
$d_{min}$ and $d_{max}$ are the minimum and maximum depth values. 
Subsequently, NeRD concatenates the learned feature at each pixel $(x,y)$ and its correspondences $(x^{\prime},y^{\prime})$ across depth $d$, resulting in a 4D feature volume $\mathcal{V}$ of size $[32 \times D \times H \times W]$. $\mathcal{V}$ is further downsampled by 3D convolutions and then flattened into a 1D vector of size $16384$ for further classification.

\medskip\noindent\textbf{(iii) Classification on the hemisphere.}
To identify the optimal symmetry plane on the hemisphere, NeRD samples a number of candidate planes $\psi$ on the hemisphere hierarchically, and then employs huge fully connected layers $(\geq 8M$ parameters) to predict the likelihood of a candidate being the symmetry plane. The sampling is repeated 3 times in a coarse-to-fine manner, until the desired resolution is reached.

\section{Data-efficient and fast 3D mirror symmetry detection}
\begin{figure}[t]
    \centering
    \includegraphics[width=1.\textwidth]{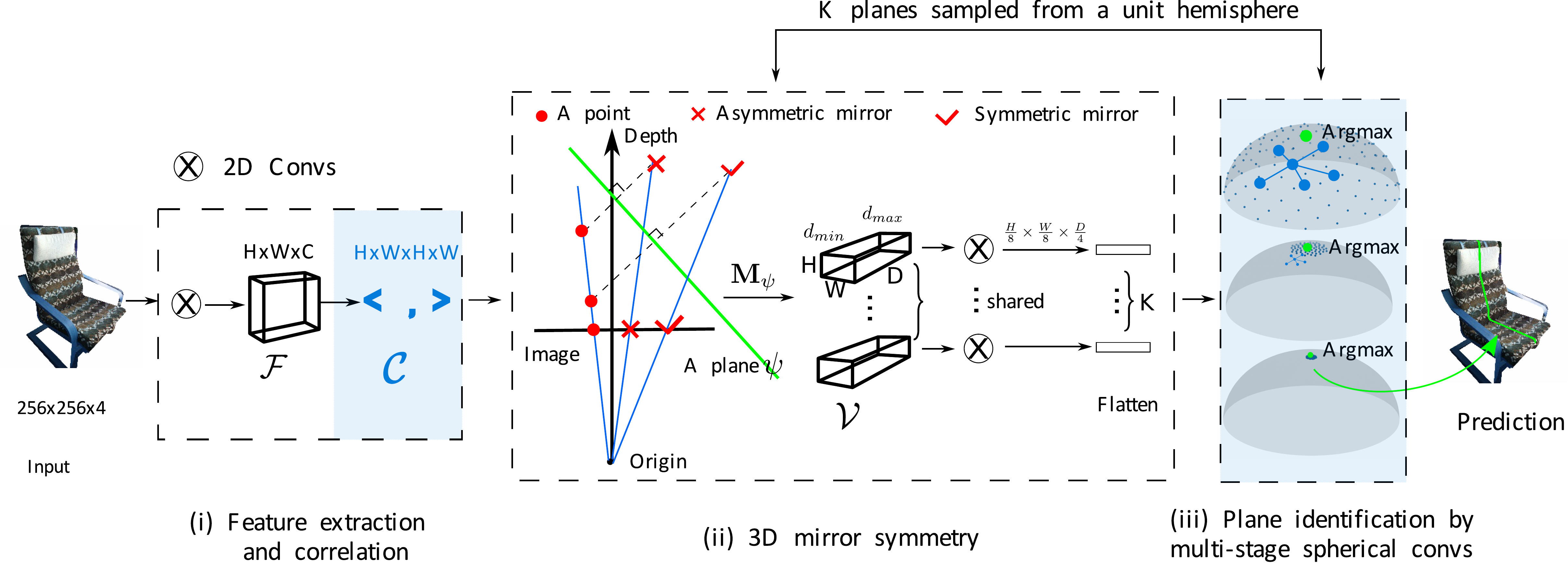}
    \caption{\small
    \textbf{Overview.} Our model follows \cite{zhou2021nerd} and includes three components: 
    (i) feature extraction and correlation calculation, (ii) 3D mirror, and (iii) plane identification by spherical convolutions. 
    The model first calculates intra-pixel correlations $\mathcal{C}$ using learned features $\mathcal{F}$. 
    Then builds a 3D correlation volume $\mathcal{V}$ for each sampled plane $\psi$, which is then flattened as a feature descriptor. 
    Here we incorporate explicit correlations to reduce computations compared to \cite{zhou2021nerd}.
    We additionally, adopt spherical convolutions on uniformly sampled planes on the hemisphere to locate the optimal plane (depicted in green).
    We highlight in blue our changes with respect to \cite{zhou2021nerd}.
    }
    \label{fig:overview}
\end{figure}

Built on top of NeRD \cite{zhou2021nerd}, our model also takes as input an RGBA image and outputs sampled planes and their associated confidence for being a true mirror plane. 
We choose the plane with the highest confidence as our prediction. 
Although an object may admit multiple symmetries, we only predict the principal mirror symmetry. 
We follow the design of NeRD \cite{zhou2021nerd}, and decompose our model into three parts, as shown in \fig{overview}: 
(i) feature extraction and correlation calculation, 
(ii) 3D mirror, 
(iii) plane identification by spherical convolutions. 
The 3D mirror follows the original design in \cite{zhou2021nerd}, while the other two parts differ, as we make two essential modifications to improve both data and compute efficiency. 
We detail these changes below.

\subsection{Compact correlation volume}
Given the learned semantic features $\mathcal{F}$ of size [$H \times W \times C$], where $H$, $W$, $C$ indicate height, width, and number of channels, we calculate intra-pixel correlations.
We correlate all pairs of points in the [$H \times W$] grid with each other by a dot product over the channel dimension.
This produces a correlation tensor $\mathcal{C}$ of size [$H \times W \times H \times W$]. $\mathcal{C}$ encodes the extent to which a pixel resembles the others. 
Using \eq{symmetry2}, we index the correlation tensor $\mathcal{C}$ at $\mathcal{C}(x,y,x^{\prime},y^{\prime})$ via bi-linear interpolation, where $(x^{\prime},y^{\prime})$ are the correspondences of a point $(x,y)$ across a candidate symmetry plane $\psi$, at various depths $d$. 
This results in a compact 3D correlation volume $\mathcal{V}$ of size [$ D \times H \times W$], which substantially reduces the computations compared to the 4D feature volume in NeRD \cite{zhou2021nerd}.
We aggregate the information over the entire $\mathcal{V}$, and apply 3D convolutions to downscale $\mathcal{V}$, resulting in an output tensor of size $\left[\frac{H}{8} \times \frac{W}{8}\times \frac{D}{4}\right]$. 
In practice, we set $H$, $W$ and $C$ to 64. 


The correlation volume $\mathcal{V}$ encodes the similarity between each input and its mirrors at all sampled depths.
A higher similarity indicates that the given plane $\psi$ is more likely to be a mirror plane. 
We flatten the downscaled volume $\mathcal{V}$ to a 1D vector.
Each sampled plane on the hemisphere $\psi$ is characterized by one such 1D vector.
\subsection{Spherical convolutions for symmetry plane detection}
Each candidate plane is equivalent to a sampled spherical point, and has an associated 1D feature descriptor obtained from the correlation volume $\mathcal{V}$. 
Given that the planes lie on the hemisphere, we take advantage of spherical convolutions to learn the most probable candidate plane. 
We use EdgeConv \cite{wang2019dynamic} to extract features from the local neighborhood of a plane.
Specifically, we treat each sampled spherical point as a node, and compute its top 16 nearest neighbors.
We stack 3 layers of EdgeConv, followed by BatchNorm and LeakyReLU activations, to guarantee a sufficiently large receptive filed. 

Following \cite{zhou2021nerd}, we also sample the planes in a multi-stage sequence by adopting the Fibonacci lattice \cite{gonzalez2010measurement}.
In practice, we sample planes over 3 stages, at multiple scales, in a coarse-to-fine manner.
$\mathbb{P}_{i} = \{\mathbf{n}_{i}^{k}\}_{k=1}^{K} \subset \mathbb{R}^{3}$ represents all sampled planes at $i^{th}$ stage, where $\mathbf{n}$ is the normal direction, and $K$ is the number of planes. The sampling at $i^{th}$ stage satisfies $\mathbb{P}_{i} = \{\mathbf{n}^{k}: \arccos(\lvert\langle\mathbf{n}^{k}, \hat{\mathbf{n}}_{i-1}\rangle\rvert)\le \delta_{i} \}_{k=1}^{K}$, where $\hat{\mathbf{n}}_{i-1}$ is the optimal plane from previous stage and $\delta$ controls the sampling region. 

During training, we minimize the binary cross-entropy loss at each stage and averaged over the positive and negative samples separately, due to the class imbalance. 
The positive samples are the spherical points that are closet to the ground truth at each stage, while the others are considered negative samples.
At test time, we choose the plane with the highest estimated confidence at the final stage.

\vspace{-2mm}
\section{Experimental analysis}
\vspace{-1mm}

\smallskip\noindent\textbf{Datasets.} 
We conduct experiments on synthetic ShapeNet \cite{chang2015shapenet} and real-world Pix3D \cite{sun2018pix3d} datasets. For both datasets the objects are aligned to the canonical space such that the Y-Z plane is the 3D mirror symmetry plane.  
On the synthetic ShapeNet dataset, we use the same subset as in \cite{zhou2021nerd} for fair comparison. 
Images are of size $256x256$ px and split in  175,122/500/8,756 training/validation/test sets. 
For the real-world Pix3D dataset we pre-process the data as in \cite{zhou2021nerd}.
We first crop the objects inside bounding boxes. 
And then, we rescale them to $256x256$ px, and adjust the camera intrinsic matrix $\mathbf{K}$ accordingly. 
This results in a dataset of 5,285 and 588 images for training and test respectively. 

\smallskip\noindent\textbf{Evaluation.}
We follow \cite{zhou2021nerd} and evaluate all methods by measuring the angle difference of the plane normals between the ground-truth and predictions in the camera space. 
We calculate the percentage of the predictions that have a smaller angle difference than a given threshold and compute the area under the angle accuracy (AA) curves.

\smallskip\noindent\textbf{Implementation details.}
The $x=0$ plane in the object space is the ground truth as it is explicitly aligned for each object \cite{chang2015shapenet}. 
We set $d_{min}=0.64$, $d_{max}=1.23$, and $D=64$ for depth. 
We perform spherical convolutions at 3 scales and sample $\{128, 64, 64\}$ symmetry planes at each scale. 
We set the scale factor to be $\delta =\{90.0^{\circ}, 12.86^{\circ}, 3.28^{\circ} \}$.  
We train from scratch for each dataset on Nvidia RTX2080 GPUs with the Adam optimizer \cite{kingma2014adam}, for maximum 32 epochs. 
The learning rate and weight decay are set to be $3 \times 10^{-4}$ and $1 \times 10^{-7}$. We decay the learning rate by 10 after 24 epochs. 
To maximize the GPU usage, we set the batch size to 6. 

\smallskip\noindent\textbf{Baselines.}
We primarily compare with the state-of-the-art: NeRD \cite{zhou2021nerd}. 
We also implement a standard baseline using direct regression to estimate the symmetry normal $\mathbf{n}$. 
For this baseline we use a ResNet-50 \cite{he2016deep} backbone with an $L_1$ loss. We additionally compare with Front2Back \cite{yao2020front2back}, which detects 3D mirror symmetry using a variant of the iterative closest point approach. 
However, Front2Back requires prior depth maps, and has only been tested on ShapeNet.
We also compare with RotCon \cite{zhou2019continuity} which proposes a continuous representation for estimating 3D rotation. 
We use the $L_1$ loss for training and report its performance on both datasets. 
We further consider DISN \cite{xu2019disn} and NCOS \cite{wang2019normalized}.
DISN learns 6D rotation representation for estimating camera poses on ShapeNet.
We recover the normal of the mirror plane from camera poses and report the performance of their pre-trained models on ShapeNet.
NCOS defines a \emph{normalized object coordinate space (NOCS)} and identifies 6D representations of camera poses. 
We use NOCS to estimate object orientation in ShapeNet.

\subsection{Exp 1: Data efficiency}

\begin{figure}
\centering
\begin{tabular}{cc}
    \begin{tabular}{c}
        \includegraphics[width=.4\textwidth]{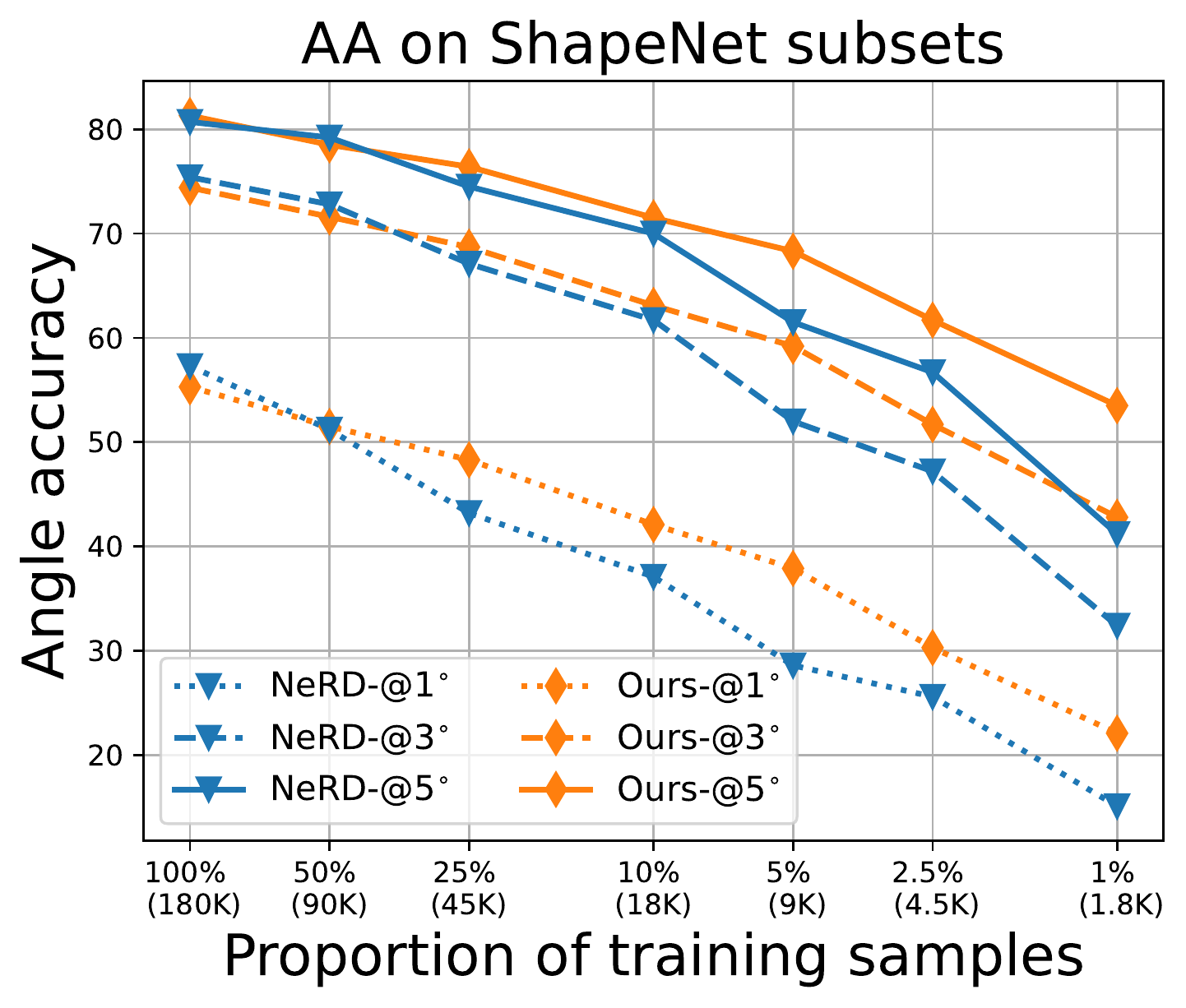} 
    \end{tabular} &
        \begin{tabular}{c}
        \includegraphics[width=.4\textwidth]{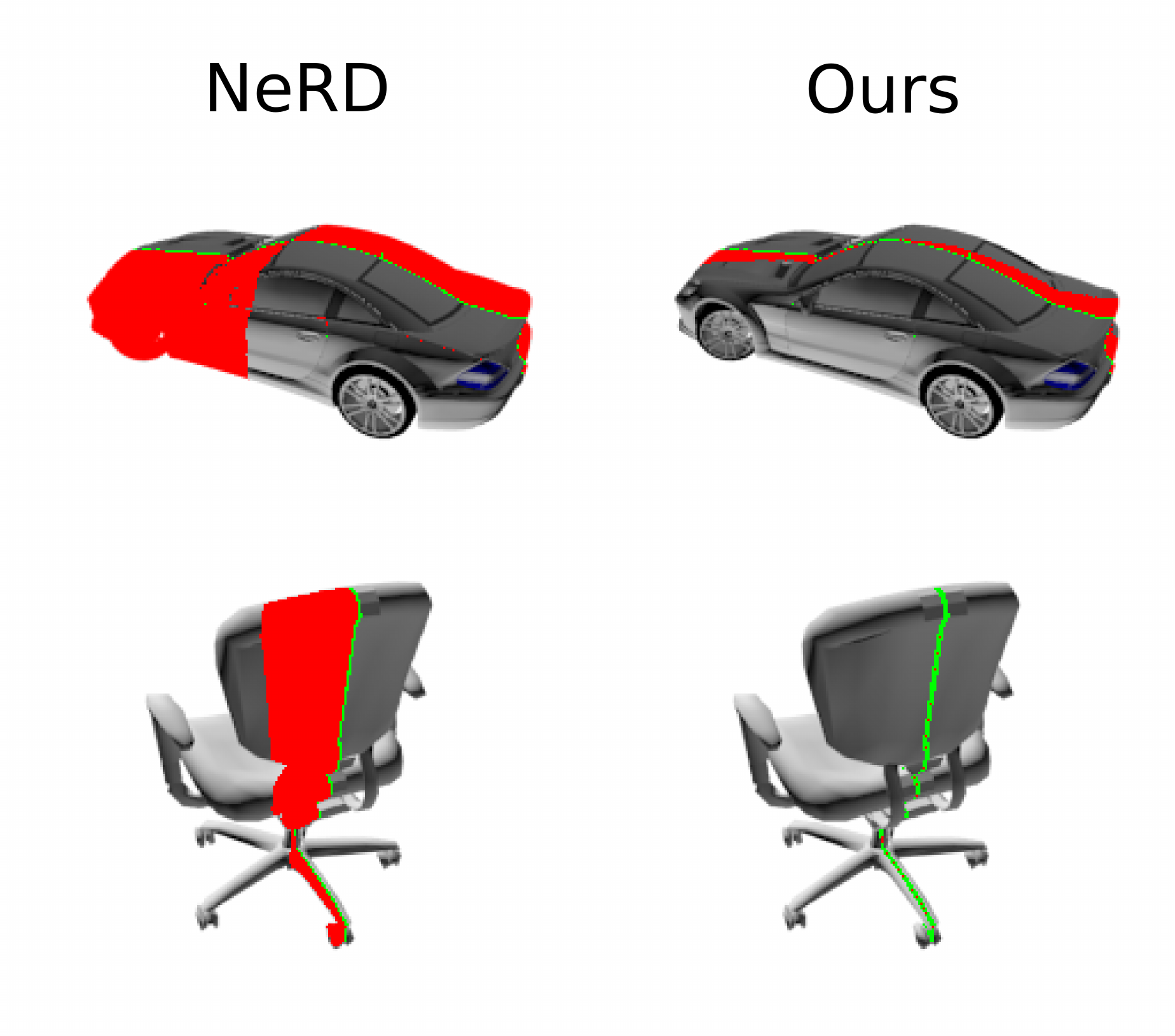} 
    \end{tabular} \\
    \small{(a) Angle accuracy on ShapeNet subsets.} & \small{(b) Examples of prediction errors.}
\end{tabular}
\caption{\small
    \textbf{Exp 1: Data efficiency.} 
    (a) Quantitative comparison between our model and NeRD when training on various ShapeNet subsets. 
    The difference of the two models is limited when the training data is ample (e.g., $18K\ge$).
    However, our model outperforms NeRD when training on limited data (e.g. $\le 9K$ images).
    (b) Qualitative comparison on the $1\%$ subset. We project detected symmetries on the image plane, where green is the true symmetry plane, and red indicates the prediction error. 
    We use the ground truth scale for plotting, to resolve the scale ambiguity. 
    Our model makes preciser predictions than NeRD when training on the $1\%$ subset.
}
\label{fig:data_efficiency}
\end{figure}

We evaluate the data efficiency of our model by reducing the number of training samples to $\{50\%, 25\%, 10\%, 5\%, 2.5\%, 1\%\}$ on the ShapeNet dataset, which has approximately 200K training images in total. 
We train all models from scratch and compare the AA scores at $3^{\circ}$ and $5^{\circ}$ on the complete test set. 
We compare our model with NeRD which holds state-of-the-art result in \fig{data_efficiency}(a), and display a few examples of detected symmetry planes in \fig{data_efficiency}(b).
The two models have a comparable amount of parameters, thus removing the impact of parameters. 
In general, our model shows superiority over NeRD with the decrease of training samples, and this advantage accentuates on subsets with fewer than 10K images. 
When training on $100\%-10\%$ subsets we observe minimal differences, while on $10\%-1\%$ subsets we observe a drastic difference (up to $10\%$ in AA).
This indicates that our model is more effective at learning from limited data. 
Notably, our model can achieve similar results to NeRD with only half of the training data, as seen on the $2.5\% $ and $5\%$ subsets, thus demonstrating the data efficiency of our model.

\subsection{Exp 2: Comparison with state-of-the art}
\begin{figure}[t]
\centering
\begin{tabular}{cc}
    \begin{tabular}{c}
            \includegraphics[width=.4\textwidth]{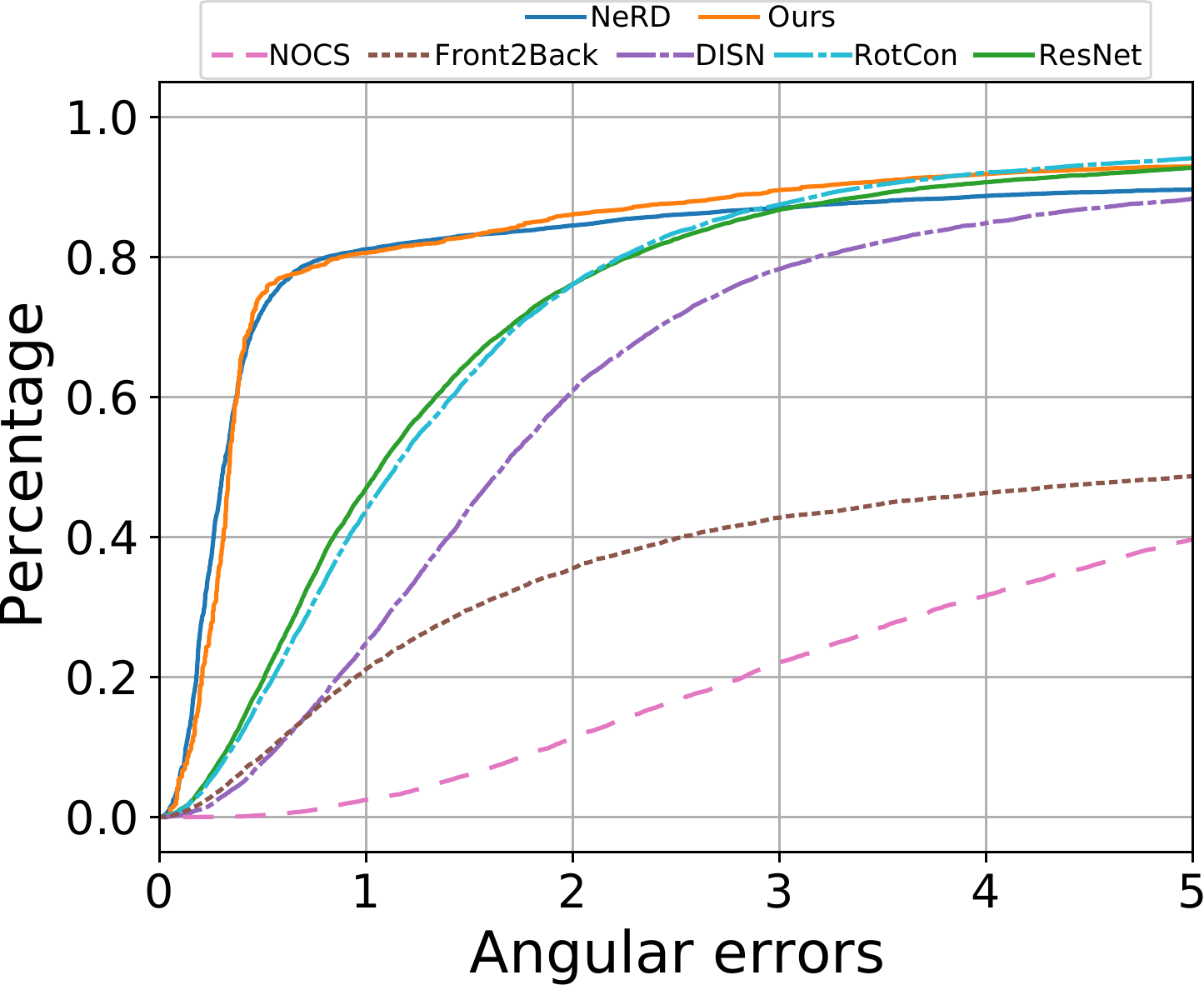} 
    \end{tabular} &
    \resizebox{0.4\linewidth}{!}{
    \begin{tabular}{p{6.5em} c@{\hskip 0.1in}ccc}
            \toprule
            Datasets  & \multicolumn{3}{c}{ShapeNet\cite{chang2015shapenet}} \\ 
            \cmidrule(l){2-4}
            Metrics  & AA$@1^{\circ}$ & AA$@3^{\circ}$ &  AA$@5^{\circ}$  \\  \midrule
            ResNet \cite{he2016deep}   & 20.8  & 55.7 & 69.5  \\
            RotCon \cite{zhou2019continuity} & 18.7  & 54.6 & 69.4  \\
            DISN \cite{xu2019disn}  & 9.3 & 26.1  & 34.1 \\
            NOCS \cite{wang2019normalized}  & 0.6 & 7.9 & 17.3 \\
            \small{Front2Back \cite{yao2020front2back}}  & 9.3 & 26.1 & 34.1 \\
            NeRD \cite{zhou2021nerd}  & \textbf{57.3} & 75.4 & 80.7 \\
            \emph{Ours}  & 55.7 &\textbf{75.5} & \textbf{82.0} \\
            \bottomrule
    \end{tabular}}\\
    \small{(a) Percentage of predictions per angular error.} & \small{(b) Angle accuracy.}
\end{tabular} 
    \caption{\small
    \textbf{Exp 2.1: Comparison on synthetic ShapeNet.} 
    Comparison with existing baselines ResNet \cite{he2016deep}, RotCon \cite{zhou2019continuity}, DISN \cite{xu2019disn}, NOCS \cite{wang2019normalized}, Front2Back \cite{yao2020front2back} and the recent NeRD \cite{zhou2021nerd}. 
    Our model shows competitive results with the top-performing model. 
    }
    \label{fig:exp_shapenet}
\end{figure}
\begin{figure}[t]
\centering
\begin{tabular}{cc}
    \begin{tabular}{c}
        \includegraphics[width=.35\textwidth]{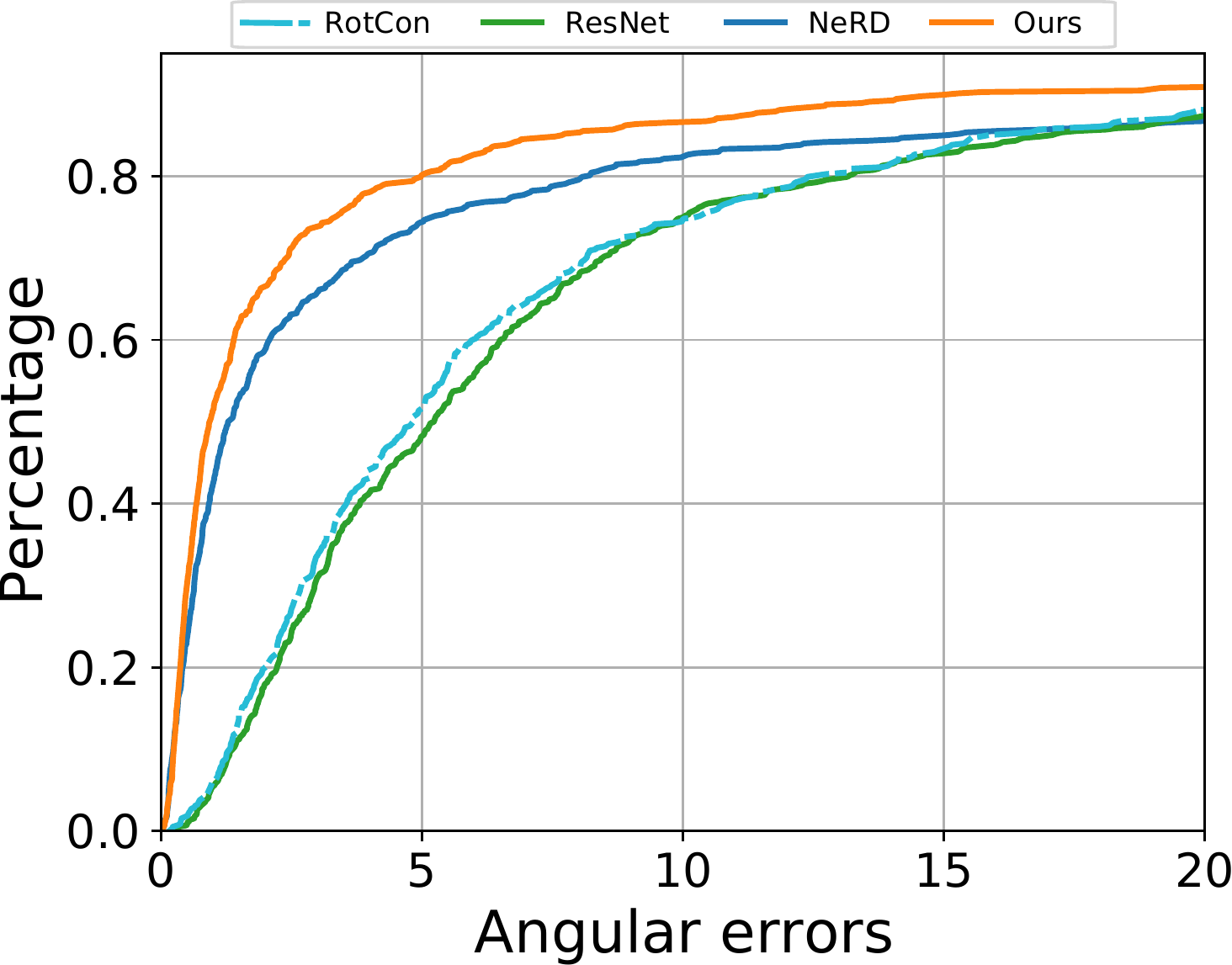}
    \end{tabular} &
    \resizebox{0.45\linewidth}{!}{
    \begin{tabular}{p{5.0em} cccc }
        \toprule
        Datasets &  Speed & \multicolumn{3}{c}{Pix3D\cite{sun2018pix3d}} \\ 
        \cmidrule(l){3-5}
        Metrics  & (FPS) & AA$@1^{\circ}$ & AA$@5^{\circ}$ & AA$@10^{\circ}$   \\  \midrule
        ResNet \cite{he2016deep} & 66 & 1.5   & 12.3  & 23.6  \\
        Rotcon \cite{zhou2019continuity}  & - & 2.2 & 14.1  & 25.8  \\
        NeRD \cite{zhou2021nerd} & 1.4 & 22.7 & 46.0 & 55.8 \\
        \emph{Ours} & 25 & \textbf{27.5}  & \textbf{53.0} &\textbf{62.8} \\
        \bottomrule
    \end{tabular}} \\
    \small{(a) Percentage of predictions                                                                                                                                                                                                                               per angular error.} & \small{(b) Angle accuracy.}
\end{tabular} 
\caption{
\small
\textbf{Exp 2.2: Comparison on real-world Pix3D dataset.} 
Our model performs the best on this challenging real-world dataset, as the knowledge of 3D mirror no longer needs to be learned from massive data. 
Moreover our model is $\times$20 faster than NeRD at inference time.
}
\label{fig:exp_pixel3d}
\end{figure}

\medskip\noindent\textbf{Exp 2.1: Comparison on synthetic data.}
In \fig{exp_shapenet} we compare with the state-of-the-art on ShapeNet. 
NeRD~\cite{zhou2021nerd} has the best AA at $1^\circ$, when large training sets are available.
Our model is competitive to NeRD. 
The prediction error of both is less than $1^{\circ}$ in $80\%$ of the test cases. 
The ResNet baseline using direct regression can only reach approximately $50\%$ AA at $1^{\circ}$, indicating that naive convolutions lack the ability to exploit the mirror symmetry, even with ample training data. 
We also notice that end-to-end approaches outperform models relying on heavy post-processing, such as Front2Back \cite{yao2020front2back}. 

\medskip\noindent\textbf{Exp 2.2: Comparison on real-world data.}
To further validate the effectiveness of our model, we also test on the real-world Pix3D dataset \cite{sun2018pix3d}, as shown in \fig{exp_pixel3d}. 
It is worth noting that the prediction error on Pix3D is relatively larger than on ShapeNet. 
On one hand, there is limited training data: $5,000$ images in total, which is significantly less than ShapeNet. 
On the other hand, the camera configuration differs from image to image, thus making it hard to make precise predictions. 
Our model outperforms all the other models consistently, thus demonstrates the superiority of our design. 
NeRD lags behind due to a high demand for training data.
Moreover, our mode is $\times$20 faster than NeRD at inference time, as showing in \fig{exp_pixel3d}(b). 
Please see the supplementary material for qualitative results.

\subsection{Exp 3: Ablation studies}
To verify the contribution of each component in our design, we conduct ablation studies, as shown in \tab{ablation}. 
All models are trained on the ShapeNet $1\%$ subset. 
Model (a) is a simple baseline using direct regression and shows inferior results to the others in detecting 3D mirror geometry. 
We replace the spherical convolutions in our design (d) with $1\times1$ convolutions in model (b). 
Comparing (b) and (d), we find that spherical convolutions improve the results significantly. 
In model (c), we replace the convolutions over the 3D cost volumes $\mathcal{V}$ by taking the max over the depth dimension. 
By doing so, we only obtain the correspondence with the highest correlation across different depths for each pixel, thus removing the 3D spatial information. 
However, model (c) substantially underperforms model (d), thus validating the necessity of 3D cost volumes. 
The ablation studies justify the added value of the 3D mirror, correlation volumes, and spherical convolutions. 

\begin{table}[t!]
    \centering
    \resizebox{0.70\linewidth}{!}{
    \begin{tabular}{c p{6.0em} p{4.0em} p{5.5em} c@{\hskip 0.1in}ccc }
    \toprule
  &  3D Mirror & Correlation volumes & Spherical convolutions & AA$@1^{\circ}$ & AA$@5^{\circ}$   \\  \midrule
  a  & \xmark &  \xmark & \xmark  & 0.8  & 9.5 \\
  b  & \cmark &  \cmark & \xmark  & 15.8  & 44.4   \\
  c  & \cmark &  \xmark & \cmark  & 8.0  & 31.5   \\
  d  & \cmark & \cmark  & \cmark & \textbf{22.1} & \textbf{53.5}  \\
    \bottomrule
    \end{tabular}}
    \caption{\small
    \textbf{Exp 3: Ablation studies.} 
    We quantitatively verify the added value of 3D mirror geometry, 3D correlation volumes, and spherical convolutions on the ShapeNet $1\%$ subset. 
    All these design choices are essential for the performance of our model.
    }
    \label{tab:ablation}
\end{table}



\section{Conclusions and drawbacks}
This paper analyzes improvements for 3D mirror symmetry detection from single-view perspective images. 
We explicitly incorporate feature correlations and spherical convolutions into the state-of-the-art 3D mirror detection \cite{zhou2021nerd}. 
This provides the model with improved data efficiency and computation efficiency. 
Extensive experiments on both synthetic and real-world datasets demonstrate the benefits of our proposed changes when compared to state-of-the-art. 

One of the drawbacks of this work is that if the found point correspondences are not sufficiently similar in appearance, the mirror plane detection will be erroneous.  
A second drawback is the incapability in detecting multiple symmetry planes, and this is restrictive as certain objects may display multiple symmetries, such as local symmetries, translational and rotational symmetries.
In addition, mirror symmetries sometimes cannot be characterized by a single plane, such as intrinsic symmetries commonly seen in non-rigid deformable objects: e.g. human bodies. 
Extending the current work for detecting multiple types of symmetries, such as rotation symmetry and translation symmetry is a viable future direction. 
Another promising future research direction is exploring the usage of 3D mirror in single-view 3D reconstruction, such as  weakly-supervised depth estimation and shape completion.

\bibliography{egbib}
\end{document}